\documentclass[journal,twoside,web]{ieeecolor}
\usepackage{tmi}
\usepackage{multirow}
\usepackage{cite}
\usepackage{booktabs}
\usepackage{amsmath,amssymb,amsfonts}
\usepackage{algorithm ic}
\usepackage{graphicx}
\usepackage{makecell}
\usepackage{xspace}
\usepackage{bm}

\DeclareUnicodeCharacter{FB01}{fi}

\usepackage{textcomp}
\def\BibTeX{{\rm B\kern-.05em{\sc i\kern-.025em b}\kern-.08em
    T\kern-.1667em\lower.7ex\hbox{E}\kern-.125emX}}
\begin{document}
\title{Reading Radiology Imaging Like The Radiologist}
\author{Yuhao Wang
\thanks{This paragraph of the first footnote will contain the date on which
you submitted your paper for review. It will also contain support information,
including sponsor and financial support acknowledgment. For example, 
``This work was supported in part by the U.S. Department of Commerce under Grant BS123456.'' }
\thanks{Yuhao Wang are with State Key Laboratory of Networking and Switching Technology, Beijing University of Posts and Telecommunications, Beijing 100876, China (e-mail:wangyuhao@bupt.edu.cn). }
\thanks{}
\thanks{}}

\maketitle

\begin{abstract}

Automated radiology report generation aims to generate radiology reports that contain rich, fine-grained descriptions of radiology imaging. Compared with image captioning in the natural image domain, medical images are very similar to each other, with only minor differences in the occurrence of diseases. Given the importance of these minor differences in the radiology report, it is crucial to encourage the model to focus more on the subtle regions of disease occurrence. Secondly, the problem of visual and textual data biases is serious. Not only do normal cases make up the majority of the dataset, but sentences describing areas with pathological changes also constitute only a small part of the paragraph. Lastly, generating medical image reports involves the challenge of long text generation, which requires more expertise and empirical training in medical knowledge. As a result, the difficulty of generating such reports is increased.

To address these challenges, we propose a disease-oriented retrieval framework that utilizes similar reports as prior knowledge references. We design a factual consistency captioning generator to generate more accurate and factually consistent disease descriptions. Our framework can find most similar reports for a given disease from the CXR database by retrieving a disease-oriented mask consisting of the position and morphological characteristics. By referencing the disease-oriented similar report and the visual features, the factual consistency model can generate a more accurate radiology report. Our model mimics the thinking process of a radiologist by utilizing both visual features and past experience with radiology imaging. Experimental results illustrate that our model achieved state-of-the-art performance on two benchmark datasets, including the IU X-Ray and MIMIC-CXR. Furthermore, the ablation study demonstrated the effectiveness of each component we proposed.
\end{abstract}

\section{Introduction}
\label{sec:introduction}
\IEEEPARstart{R}{adiology} 
Automated radiology report generation aims to generate comprehensive and accurate reports that contain abundant abnormal observations about medical images. Writing such reports manually is time-consuming and difficult, requiring the expertise of an experienced radiologist. Fully automated report generation can assist radiologists in writing imaging reports, improving the accuracy of disease detection, and reducing their workload. Additionally, automated radiology report generation can provide automatic medical image reports to regions with limited access to medical resources, helping to alleviate the shortage of local experts. Unlike traditional healthcare AI tasks like disease classification, radiology reporting requires AI models to possess a higher level of cognitive ability to produce medical image reports with descriptive expressions that resemble human-level cognition.

Medical reports are generated based on the task of image captioning. Considerable progress has been made in research on image captioning, with most frameworks adopting an encoder-decoder architecture, such as a CNN image encoder followed by an RNN decoder for report generation. The Transformer architecture, initially proposed for text modeling and later extended to visual-language tasks, has been widely applied in cross-modal domains due to its effectiveness in modeling sequences and reducing the semantic gap between images and text using stacked encoders and decoders with multi-head self-attention. Recent research in image captioning predominantly adopts the Transformer architecture as the main architecture.

However, radiology report generation differs significantly from image captioning. The main differences can be summarized as follows: 1. Most radiology images are highly similar to each other, with only subtle differences in the areas with pathological changes. Existing models often struggle to attend to the specific regions of these lesions, making it challenging to generate descriptions that focus on subtle pathological areas. 2. Medical image reporting involves the challenge of generating long-form text, which is more difficult, and the supervision signal for describing disease-specific information is often sparse. Radiology report generation also requires more expertise knowledge and training compared to image captioning in the natural image domain. 3.The data bias in radiology report generation is serious, resulting in the problem of shortcut learning. Deep learning models often generate radiology reports that lack significant disease descriptions crucial for clinical diagnosis.

To address the inherent problems in radiology report generation, we propose the RRGnet framework in this paper. It is known that for a medical imaging examination with a disease, only a small part of the corresponding imaging report describes the relevant disease. This situation causes important diagnostic terms to be buried within a large amount of normal statements. Predicting the disease tag of CXR is generally easier and more accurate compared to generating a long description of the symptoms, as it provides a stronger supervision signal. We leverage interpretable artificial intelligence techniques and class activation maps, widely used in tasks such as weakly supervised object detection or image segmentation, to reflect the location and morphological characteristics of targets. Inspired by works that apply class activation maps to analyze the decision-making process in deep learning models, we propose a disease-oriented mask retrieval module. This module effectively retrieves more accurate reports from the database at the disease perspective. Experimental results and qualitative analysis demonstrate the effectiveness of the disease-oriented mask retrieval module in finding samples with the same disease as the input sample, exhibiting a high degree of consistency in disease location and morphological characteristics.

The search module based on a disease-oriented mask finds the most similar image reports from the database as reference reports, simulating how radiologists refer back to reports they have seen in the past with the same disease. Additionally, we propose a fact-consistent image report generation module based on copying mechanism. This module utilizes vocabulary-level prior information during the generation of medical image reports in the decoder, enhancing the clinical accuracy and fact consistency of the generated reports. The attention concept within the copying mechanism simulates the varying degrees of reliance that radiologists typically have on different reference reports. Furthermore, the model focuses on the parts of the original image that differ from similar textual descriptions, integrating prior knowledge and input images to generate accurate medical image reports.

The main contribution of this paper can be summarized as follows:
\begin{itemize}
	\item We propose a CAM-based\cite{CAM} similarity retrieval method that generates disease-oriented masks, which effectively represent the disease morphology and location information. This significantly improves the accuracy of similar report retrieval, enabling the corresponding decoder to attend to a greater extent to disease-specific descriptions during the decoding stage.
	\item  We propose a fact-consistent image report generation module based on a copying mechanism. This module simulates the writing process of radiologists when composing image reports and effectively utilizes the input's relevant diseases. It greatly enhances the clinical accuracy and fact consistency of the generated medical image reports.
	\item   We conducted experiments on two widely used medical image report generation datasets, IU-Xray\cite{IU-Xray} and MIMIC-CXR\cite{MIMIC}. Both qualitative and quantitative experiments were performed to demonstrate the effectiveness of our proposed model.
\end{itemize}

\section{Related Work}

\subsection{Image Captioning}
Image captioning, which aims to generate descriptive and meaningful captions for images, has been extensively explored in deep learning. Early approaches to image captioning relied on handcrafted features and language models \cite{Deepvisualsemantic,ATT2IN,showattendandtell,bottomandup}. However, with the advent of deep learning, convolutional neural networks (CNNs) and recurrent neural networks (RNNs) have become the predominant architectures for image captioning \cite{showandtell}. These architectures allow for end-to-end training, enabling the models to learn both visual and textual representations.

Attention mechanisms have been proposed to improve image captioning by allowing the model to focus on relevant image regions during the caption generation process \cite{attention}. By attending to specific regions, the model can align the generated words with the corresponding visual content, resulting in more accurate and contextually relevant captions \cite{KnowingWTolook,M2TR,meshedmemory}. Furthermore, reinforcement learning techniques have been applied to optimize captioning models by incorporating reward-based feedback \cite{reinforce}. This approach involves training the model to maximize a reward signal, typically based on the quality of the generated captions, which can lead to improved captioning performance through iterative optimization.

\subsection{Image Retrieval}
Image retrieval\cite{Imageretrievalsurvey} is the task of retrieving similar images from a database based on the content of the images. 
One approach is to fine-tune convolutional neural networks (CNNs) with a ranking loss function \cite{deepimageretrieval,retrievalcnn1}. By optimizing the network parameters based on a ranking loss, the model learns to better differentiate between relevant and irrelevant images, resulting in improved retrieval performance. Attention mechanisms have also been incorporated into the image retrieval process \cite{deepimageretrieval2,transformerimageretrieval}. Inspired by human visual perception, attention mechanisms allow the model to focus on relevant parts of the image, improving the ability to capture and match distinctive features for retrieval. Despite these successes, challenges persist in deep learning-based image retrieval. One major challenge is the semantic gap that exists between low-level visual features and high-level semantics. Low-level features extracted from images may not capture the complex semantic meaning, making it difficult to accurately match images based on their content.

\subsection{Radiology Report Generation}
Radiology report generation is similar to image captioning in many ways, but there are some notable differences. While image captioning typically generates single-sentence captions, radiology reports are paragraphs containing multiple sentences. To address this difference, some methods have adapted hierarchical LSTM models \cite{automatic,clinicallaccurateCXR,ShowDA,Whenradiology} to handle the generation of longer radiology reports. In particular, the work by \cite{automatic} employed chest X-ray disease classification as an auxiliary task to improve automatic radiology report generation.

Some radiology report generation methods\cite{KERP,8HybridRR,yang2022knowledge,EDPPK} utilize image retrieval to obtain a template report that closely resembles the input image. However, these methods suffer from two major shortcomings. Firstly, due to the similarity of chest X-ray images, it is difficult for these methods to retrieve similar reports for images with similar diseases. Secondly, these methods represent the retrieved similar report as a prior knowledge vector, which limits the model's ability to leverage the rich linguistic properties of the reports.

With the development of attention mechanisms \cite{attention}, transformer models have emerged as powerful tools for bridging the gap between image and text modalities. Recent works \cite{R2Gen,R2Gencmn,aligntransformer,EDPPK,M2TR} have adopted transformer encoder-decoder architectures for radiology report generation and demonstrated excellent performance. However, most existing methods rely solely on a visual encoder trained jointly with a decoder to extract information from the image, without explicitly leveraging the linguistic properties of similar radiology reports.

\subsection{Class Activation Map}
The Class Activation Map (CAM) technique was initially proposed to highlight the regions in an image that are most important for a model's prediction \cite{CAM}. It has primarily been used in image classification tasks \cite{CAM,GradCAM}, where the objective is to identify the main objects in an image. CAM-based methods have been employed in weakly supervised object detection \cite{CAMobjectDet1,CAMobjectDet2,CAMobjectDet3} to improve localization capabilities when only image-level annotations are available. Additionally, weakly supervised semantic segmentation works \cite{CAMSeg1,CAMSeg2,CAMSeg3} have utilized CAM to generate pseudo segmentation labels and train segmentation models.

The success of CAM-based approaches in weakly supervised object-level tasks highlights their ability to capture position information and morphological characteristics of objects. Inspired by these works, we apply CAM to generate disease-oriented masks and incorporate them into the image retrieval process. By leveraging CAM, we aim to retrieve more accurate and relevant reference reports for similar diseases. To the best of our knowledge, this is the first paper to employ the CAM method to advance the task of radiology report generation.

\begin{figure*}
    \centering
    \includegraphics[width=0.95\textwidth]{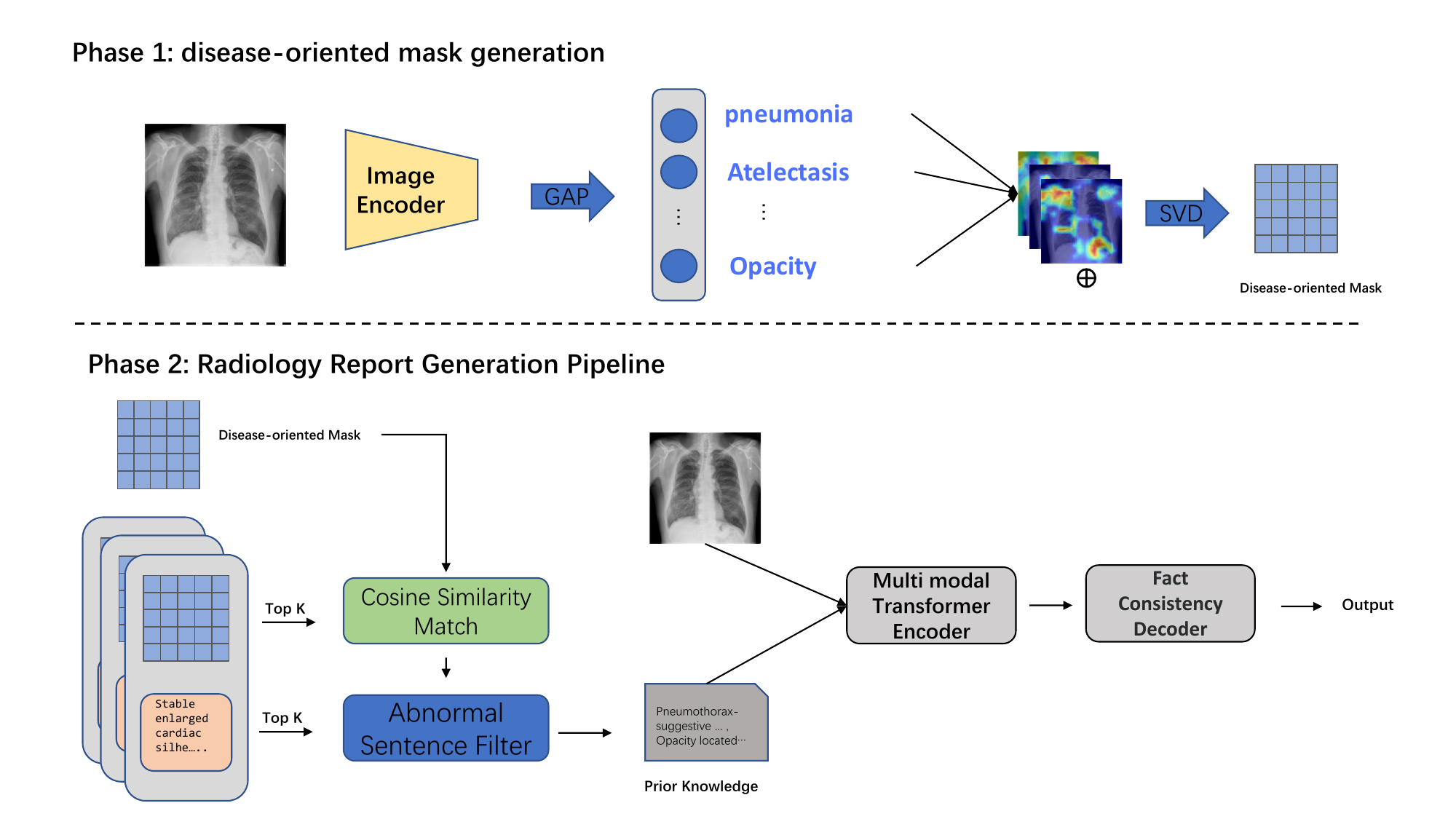}
    \caption{The model consists of two stages: disease-oriented mask generation based on CAM and fact-consistent medical image report generation based on the copy mechanism. In the first stage, a disease-oriented mask, which can represent rich disease information, is generated for all images using the CAM mechanism. In the second stage, the disease-oriented mask is used instead of the original image for similarity retrieval, in order to obtain medical image reports that are highly similar to the input image in terms of diseases as prior knowledge. Then, a fact-consistent decoder based on the copy mechanism synthesizes the corresponding medical image report based on different attention weights for the image and vocab-level prior knowledge.}
    \label{figure1}
\end{figure*}

\begin{figure*}
    \centering
    \includegraphics[width=0.8\textwidth]{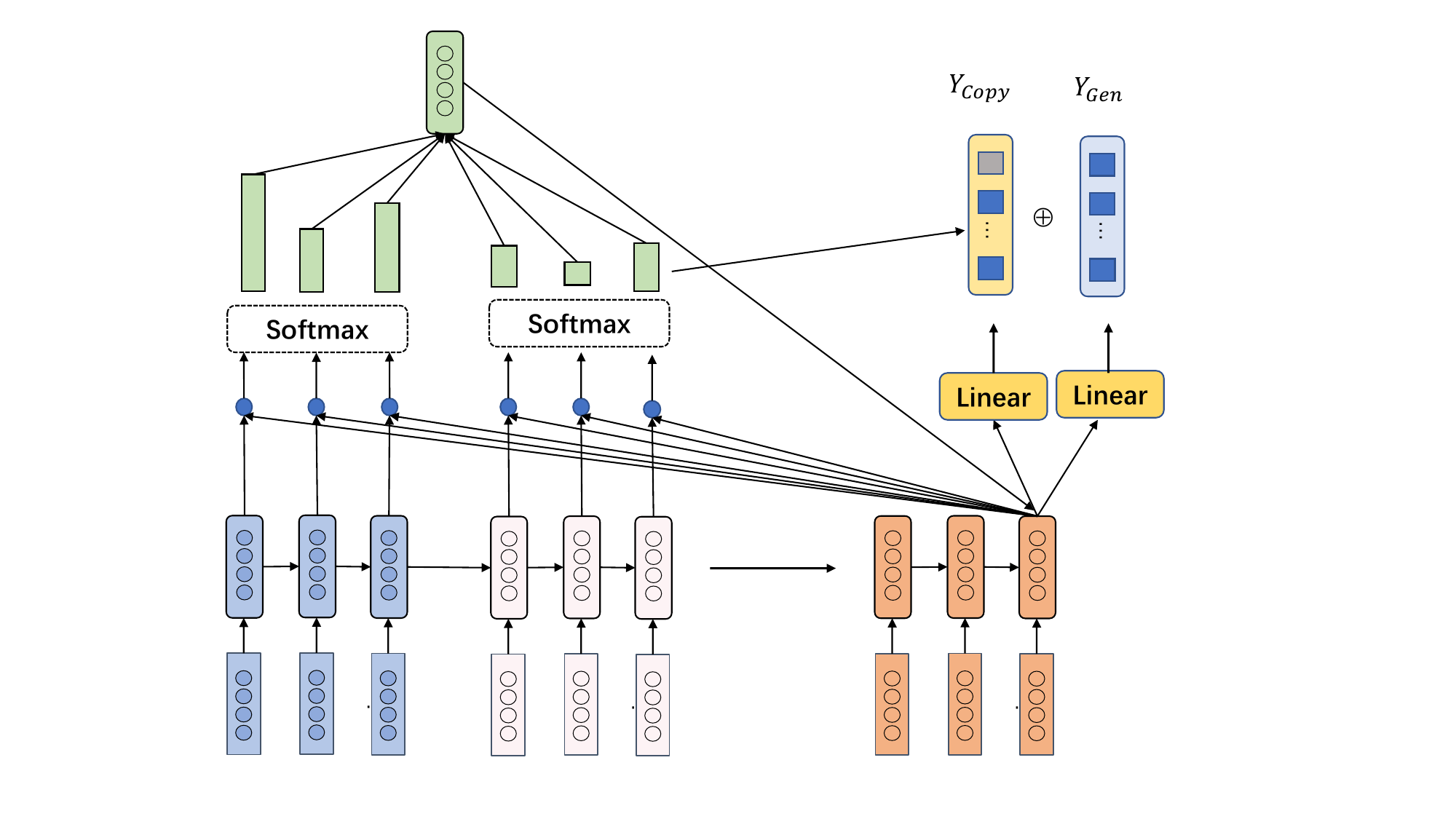}
    \caption{The illustration of fact-consistent decoder}
    \label{figure4}
\end{figure*}

\section{METHODOLOGY}

The proposed method is illustrated in \ref{figure1}. It consists of two stages aimed at improving radiology report generation. In the first stage, we generate a disease-oriented mask using the Class Activation Map (CAM) technique \cite{CAM}. This is achieved by aggregating the class activation maps corresponding to different disease labels. The aggregated disease representation matrix is then subjected to dimensionality reduction through Singular Value Decomposition (SVD) to obtain the disease-oriented mask. The disease-oriented mask effectively captures various disease information along with their corresponding morphological and location details. This enables precise retrieval of similar diseases during the retrieval process, resulting in higher-quality corresponding image reports as prior knowledge. In the second stage, we incorporate the copy mechanism and propose a fact-consistent image report generation model. The decoder of the model takes advantage of both prior knowledge and input images, allowing for comprehensive consideration of both sources of information during the generation of text tokens. By leveraging prior knowledge and images together, the model generates more clinically meaningful and efficient image reports.

\subsection{Phase 1:Disease Oriented Mask Generation}

The process of generating a disease-oriented mask simulates the procedure used by radiology experts to analyze diseases. Initially, we extracted disease labels from the corresponding medical imaging reports using Chexbert\cite{chexbert}. These disease labels encompass a range of common diseases such as 'Enlarged Cardiomediastinum,' 'Cardiomegaly,' 'Lung Opacity,' 'Lung Lesion,' 'Edema,' 'Consolidation,' 'Pneumonia,' 'Atelectasis,' 'Pneumothorax,' 'Pleural Effusion,' 'Pleural Other,' 'Fracture,' and 'Support Devices.' These extracted labels encompass a comprehensive set of diseases.

Next, we employed a CNN network with a global average pooling layer to perform multi-label classification for the aforementioned diseases. Additionally, we applied the Class Activation Mapping (CAM) method\cite{CAM}to obtain a class activation map for each disease category. Subsequently, we aggregated multiple class activation maps at the channel level. To enhance retrieval efficiency, we utilized the Singular Value Decomposition (SVD) method to reduce the dimensionality of the aggregated class activation map. This process resulted in a matrix that characterizes the location and morphology information of various diseases, which we defined as the disease-oriented mask.

For a given image $I \in \mathcal{R}^{H*W*C}$, the activation of the k unit feature maps of the last convolutional layer at the spatial location (x, y) is represented by $f_k(x, y)$ for a given image. With performing GAP layer, the class activation map of c class can be formulated as:
$$S_c=\sum_kw_k^c\sum_{x,y}f_k(x, y), S_c\in \mathcal{R}^{H*W*1}$$
where $w_k^c$  is the weight corresponding to class c for unit k. Subsequently, the disease oriented mask can be obtained by :
$$DOM_{i}=\left[S_1,S_2,.....S_K\right], DOM_{i}\in \mathcal{R}^{H*W*K} $$
k represents the number of diseases defined in the classification model. In our actual experiments, K represents 14, which is the number of disease labels that can be extracted by CheXbert\cite{chexbert}. To reduce memory usage and improve retrieval efficiency, we compressed and reduced the dimensionality of disease-oriented mask using SVD for storage. Finally, after the phase 1, we added disease-oriented mask for each image-report pair. Therefore, the basic format of the dataset can be expressed as follows: $\left(I,T,DOM)\right)$ where $I,T$ denoted the original Image and radiology report, $DOM$ is corresponding disease-oriented mask.

\subsection{Phase 2:Fact consistency based radiology report generation}

The module for generating radiology reports based on fact consistency consists of three components: 1. Similarity report retrieval module rely on disease-oriented mask 2. Prior information extraction and representation module 3. Fact-consistent image report generation module based on the copying mechanism.

\subsubsection{Similar Report Retrieval Module}
For each input data unit $\left(I_i,T_i,DOM_i\right)$, we have its corresponding disease-oriented mask. We calculate the cosine similarity to obtain the similarity scores between the disease-oriented mask and the disease-oriented mask pool. specifically,the knowledge pool is the disease-oriented mask pool: $DOM_{Pool}=\left[DOM_1,DOM_2,....DOM_N\right]$. We calculate the cosine similarity between the input $DOM_i$ and $DOM_{Pool}$, and selecting Top k samples' radiology report as reference reports.
Generally, radiology report is
expressed as $T=\{s_1,s_2,...s_l\}$,where
$s_i$ denotes the $i$ th sentence.
Meanwhile,the sentences $s_i$ is a long sequence $\{w_1,w_2,...,w_T\}$ and $w_i$ is the $i$ token of the reference report.

Using disease-oriented masks, the retrieved prior knowledge demonstrates a remarkable consistency with the input report in terms of disease localization and morphological information. Qualitative analysis has effectively showcased the efficacy of our method in retrieving reports from patients with the same disease, displaying a strong resemblance to the target report and achieving a notable alignment at the disease level. Additionally, we have observed a high degree of consistency between the position and morphological descriptions of the disease in certain retrieval results, with sentences describing abnormalities being remarkably similar. Based on these findings, we can confidently conclude that the retrieved prior knowledge is highly relevant to the target report. It serves as fact-consistent knowledge, contributing to the enhancement of medical image report quality generated by our model. Typically, after Retrieval, the basic data unit is $\left(I_i,T_i,R_1,R_2,...R_K\right)$, where $R_i$ denotes the reference report.

\subsubsection{Prior information extraction and representation module}

Due to the fact that sentences describing abnormalities in medical image reports typically occupy a small portion of the overall report, utilizing all image reports as knowledge input often leads to a significant amount of redundant information. This redundancy can hinder the model's ability to focus on crucial disease-specific details.
To address this issue, we utilized Chexbert\cite{chexbert} to analyze every sentence in all of the imaging reports. This approach can effectively identify the sentences that truly describe the diseases in an imaging report, thus obtaining more refined prior knowledge as prior knowledge.

For instance, the report "Lungs are clear. No pleural
effusions or pneumothoraces.heart size is upper limits of normal, There are low lung volumes with bronchovascular crowding and scattered opacities in the bilateral lung", after identifying, the prior knowledge are "heart size is upper limits of normal" and "There are low lung volumes with bronchovascular crowding and scattered opacities in the bilateral lung".

By extracting all possible relevant information regarding lesions, we augment the quantity of valuable information derived from the prior knowledge. Subsequently, we encode both the original image and prior knowledge to obtain multimodal representations. The textual representation preserves vocabulary-level information, empowering the decoder to generate higher-quality medical image reports using the copying mechanism. The clinical effectiveness of the generated reports has been validated through the analysis of relevant metrics.
Typically, after Retrieval, the basic data unit is $\left(I_i,T_i,R_1,s_2,...s_m\right)$, where $s_i$ denotes the abnoumal sentences, and $m$ is the number of sentences.

\subsubsection{Fact-consistent image report generation module}

Pointer Networks\cite{PointerNetworks}are specifically designed for sequential decision-making tasks, addressing scenarios where the network needs to select elements from an input sequence based on contextual information. Unlike generating discrete tokens, Pointer Networks employ attention mechanisms to directly output indices or positions within the input sequence. The architecture comprises an encoder, attention mechanism, and decoder, enabling it to handle output sequences of variable lengths. Pointer Networks have demonstrated effectiveness in tasks such as routing, combinatorial optimization, and structure parsing. Furthermore, they have been widely adopted in text summarization, as the copying mechanism effectively captures key words in the input text, enhancing the accuracy of summary extraction.
Specifically, the copying mechanism posits that the output generated by a model is derived from the input. 

In each time-step of the decoder, general seq2seq\cite{attention} model produces the vector that influences content-based attention weights corresponding to the input sequence.
In the case of the pointer network, these attention weights act as indicators pointing to specific positions in the input sequence. The input time-step with the highest weight is considered the output for that particular decoder time-step.
The formulation can be be expressed in \ref{pointerequ}

$$
u_j^i=v^T \tanh \left(W_1 e_j+W_2 d_i\right) \quad j \in(1, \ldots, n)\label{pointerequ}
$$

$$
p\left(C_i \mid C_1, \ldots, C_{i-1}, \mathcal{P}\right)=\operatorname{softmax}\left(u^i\right)
$$
the encoder and decoder hidden states as $(e_1, . . . , e_n)$ and $(d_1, . . . , d_p)$, and $v^T,W_1$ are all parameters of the model. The output of softmax operation points to the input token having the maximum value.

In the process of generating medical imaging reports, doctors also exhibit a similar implicit thinking process. When reviewing medical images, doctors draw upon their encounters with similar images in the past and reference previous writing styles while composing reports. Moreover, they need to consider the distinctions between existing imaging and reference images. Our model effectively simulates this thinking process. As the decoder produces an imaging report, it can simultaneously consider highly reliable prior knowledge and the original input image. The image report generation module, based on fact consistency, fully leverages vocabulary-level prior information while focusing on the input image, resulting in more precise medical imaging reports. Our specific implementation is as follows:

$$
p\left(\mathcal{C}^{\mathcal{P}} \mid \mathcal{P} ; \theta\right)=\prod_{i=1}^{m(\mathcal{P})} p_\theta\left(C_i \mid C_1, \ldots ,C_{i-1}; \mathcal{K}; \mathcal{I} ; \theta\right)
$$

here $\mathcal{C}^{\mathcal{P}}=\{C_1, \ldots ,C_m(\mathcal{P})\}$ is target report, consisting a 
sequence of text tokens.
 $\mathcal{I}=\{I_1, \ldots ,I_{n}\}$ is the the radiographics, are mede up of a sequence of image pathes tokens.
$\mathcal{K}=\{K_1, \ldots ,K_m\}$ is prior knowledge, which is composed of a sequence of text tokens.

Given a training triplet($\left(I_i,T_i,K_i\right)$,We denote the final output of the decoder as $z_1, ..., z_t$, our model aims at computing the conditional probability $z_i$:

3$$\mathbf{Y}\left(z_i\right)=\frac{\left(\mathbf{Y}_{Gen}\left(z_i\right)+\mathbf{Y}_{Copy}\left(z_i\right)\right)}{2}$$

The output probability $\mathbf{Y}\left(z_i\right)$is composed of both the image attention $\mathbf{Y}_{Gen}\left(z_i\right) $and the prior knowledge attention $\mathbf{Y}_{Copy}\left(z_i\right)$ in the model. Specifically, the model generates predictions by attending to both the input image and prior knowledge.

For the image attention part, we calculate the attention coefficients between the decoder hidden state vector and all the encoder input hidden state vectors. The softmax function normalize the attention coefficients $u_j^i$ over all the image patches in the input.

$$
\begin{array}{rlr}
u_j^i & =v^T \tanh \left(W_1 I_j+W_2 d_i\right) & j \in(1, \ldots, n) \\
a_j^i & =\operatorname{softmax}\left(u_j^i\right) & \\
d_i^{\prime} & =\sum_{j=1}^n a_j^i e_j &
\end{array}
$$

where $v^T,W_1,W_2$ are all learnable parameters,
$I_j$ is the element of the input image patch sequences $\mathcal{I}$. The $d_i^{\prime}$ is the generation vector,$d_i$ is the decoder hidden states. Then, the $\mathbf{Y}_{Gen}\left(z_i\right)$ is obtained by:

$$\mathbf{Y}_{Gen}\left(z_i\right)=softmax(Linear\left(d_i^{\prime};d_i\right))$$ 
Lastly, $d_i^{\prime}$ and $d_i$ are concatenated and used as the hidden states from which we make predictions and which we feed to the next time step in the recurrent model.

For the part that incorporates prior knowledge, we calculate the similarity coefficients between the decoder hidden state vector and the token embeddings of all the prior knowledge inputs. The attention coefficients indicate the relevance of the prior knowledge to the current decoding step, and are used to compute the probability of outputting each token at the current time step. Specifically, in the last transformer block, attention weights are generated that represent the probabilities of copying the text from each token of prior knowledge. We define a token $z_i$ to be produced if a node $k_j$ is selected, and the text of that token starts with $z_i$.

The computation process for incorporating prior knowledge using attention is as follows:

\textbf{$$
\begin{array}{rlr}
u_j^i & =v^T \tanh \left(W_1 K_j+W_2 d_i\right) & j \in(1, \ldots, n) \\
a_j^i & =\operatorname{softmax}\left(u_j^i\right) & \\
\end{array}
$$}

$$\mathbf{Y}_{Copy}\left(z_i\right)=\sum_{\substack{j \in V\\j=z_i}} a_j^i$$ 
In summary, our model for generating image reports considers both the token generation probability from the input image and the token copying probability from the prior knowledge, which helps improve the quality of the generated reports and their clinical relevance. The detail of the Fact-consistent image report generation module is illustated in \ref{figure4}

\section{EXPERIMENTS}

\subsection{Datasets and Tool}

\subsubsection{IU-Xray}

IU X-Ray is a widely recognized benchmark dataset for evaluating medical image report generation models. The dataset comprises over 7470 chest X-rays and 3955 corresponding radiology reports, which have been manually annotated by expert radiologists. The reports typically consist of multiple sentences, outlining the observations, impressions, and recommendations based on the images. We 
 adopt the "Findings" section which provides a detailed paragraph describing the observed evidence as the target sequence.

\subsubsection{MIMIC-CXR}
MIMIC-CXR is a dataset comprised of 64,588 patients collected at the Beth Israel Deaconess Medical Center between 2011 and 2016. The MIMIC-CXR dataset was collected over multiple years and encompasses a large volume of patient data. It provides a rich resource of chest X-ray images and associated radiology reports, enabling extensive research and algorithm development in the field of chest imaging analysis. It includes 77,110 chest X-ray images and 227,835 corresponding free-text radiology reports. To ensure experimental fairness, we followed the experimental setup of previous studies, resulting in a training set of 222,758 samples, and validation and test sets consisting of 1,808 and 3,269 samples. 
\subsubsection{Chexbert}
CheXbert is a method that combines automatic labeling and expert annotations to accurately label radiology reports using BERT. It can annotate 14 common medical observations, including fractures, consolidation, enlarged mediastinum, not detected, pleural other, cardiomegaly, pneumothorax, atelectasis, support devices, edema, pleural effusion, lung lesions, and lung opacities. We utilized CheXbert to perform label extraction on IU-Xray and MIMIC-CXR datasets, extracting corresponding disease labels. Since CheXbert was pre-trained on MIMIC-CXR, it provides more accurate disease label extraction for MIMIC-CXR. Therefore, subsequent analysis experiments, such as ablation studies, were mainly conducted based on MIMIC-CXR.

\subsection{Implementation Details}
In the first stage, we employ ResNet\cite{resnet} with GAP layer as the network for disease classification, and adopt Class Activation Maps (CAM)\cite{CAM} as the method for generating disease-oriented masks. 
The size of each class activation map for each category is 2242241. Furthermore, Chexbert provides a total of 14 disease labels. By aggregating multiple class activation maps along the channel dimension, we obtain a disease-oriented mask with dimensions of 224*224*14. During the generation of disease-oriented masks, we utilize Singular Value Decomposition (SVD) to reduce the dimensionality of the aggregated class activation maps, thereby enhancing the retrieval efficiency and reducing the storage space required for the masks. After compression, the size of each disease-oriented mask is 224*224*3.

In the second stage, we get the disease-oriented mask vectorization and computed the similarity between the disease-oriented mask of the source image and the disease-oriented masks in the mask pool, then selected the top k medical image reports that strictly cannot correspond to the source image as prior knowledge. k is a hyperparameter, and we define it as 3 in our main experiment. We used all samples in the training set to construct the disease-oriented mask pool.

Subsequently, we extract sentences from the selected reports that indicate the presence of diseases, serving as prior knowledge. Specifically, we use Spacy\cite{Spacy} to tokenize a given reference imaging report into sentences based on punctuation. After dividing the text into individual sentences, we apply Chexbert for annotation analysis. We retain the sentences that contain positive disease labels as reference prior knowledge. We feed both the original images and the prior knowledge into a multimodal input encoder. The number of layers in the multimodal input encoder and text decoder is set to 3, and the number of attention heads is set to 8.
All input images are resized to 224*224 pixels and split into 7*7 patches. We concentrate several abnormal sentences.The max token length of prior knowledge is set 100. The maximum output token length is set to 60. We utilize the Adam optimizer with a learning rate of 1e-4. The training process spans 100 epochs, while maintaining the same parameter settings for both datasets. We use two NVIDIA A40 GPUs to trained our model and set the batch size 32. The experimental settings remain consistent across the  IU-Xray and MIMIC-CXR.

\subsection{Evaluation Metric}
 
\subsubsection{Natural Language Generation Metrics}
To evaluate the generated report quality, we adopted BLUE-1,BLUE-2,BLUE-3,BLUE-4\cite{bleu},ROUGE-L\cite{rouge},METEOR\cite{meteor},CIDR\cite{cider} as the NLG metric. The assessment of predictive reports' descriptive accuracy relies on the utilization of NLG metrics. BLEU (Bilingual Evaluation Understudy) was originally designed for machine translation tasks and calculates the overlap of word n-grams between predictions and references, capturing the semantic and fluency aspects of sentences. However, it has limitations in accurately evaluating long sentences. Rouge effectively considers recall and enables effective evaluation of long texts. Meteor combines multiple evaluation metrics, taking into account precision, recall, and edit distance, among other factors. It considers the similarity of synonyms and word order, which allows for better evaluation of semantic variations. CIDER considers lexical consistency and coherence evaluation, providing a comprehensive assessment of the similarity between generated descriptions and reference descriptions. It is widely used in image captioning tasks. The CIDEr value is used to evaluate whether a model can generate more accurate thematic descriptions by assessing the frequency of non-repetitive sentences in the training set.

\subsubsection{Clinical Efficacy Metrics}

To evaluate the clinical efficacy of the generated report, we utilized Chexbert\cite{chexbert} to extract disease labels from the model-generated report. Precision, recall, and F1-score were employed as metrics to assess the clinical efficacy of our model. It should be noted that Chexbert was pretrained on MIMIC-CXR, and its extraction results may not be sufficiently accurate for IU-Xray. Therefore, we only present the clinical efficacy metrics based on the MIMIC-CXR datasets.

\section{Results and Discussion}

\subsection{Comparison with SOTA}

\subsubsection{Description Accuracy}

We compare our methods with a range of previous SOTA radiology report generation methods and image captioning methods.
For Image captioning methods, State-of-the-art (SOTA) models in the field of image captioning, such as ADAATT\cite{KnowingWTolook},ATT2IN\cite{ATT2IN},CoAT\cite{automatic}, that utilize encoder-decoder architectures, are included in the comparison.

For previous radiology report generation methods, we compared our methods with 
R2Gen\cite{R2Gen},CMN\cite{R2Gencmn} which employ the memory mechanism to restore the patter information during train process and other knowledge enhanced radiology generation methods like KERP\cite{KERP},HRGR\cite{HRGR},ARRG\cite{ARRG}
Methods like CMCL\cite{CMCL},CA\cite{CA} utilize contrastive learning to model the pairing relationship between images and text, enhancing the generation of image captions.
As show in \ref{tab:main_results},our methods achieve State-of-the-Art among all metrics comparing with all other methods. This indicates that our model has achieved comprehensive improvement in terms of language fluency and accuracy in generating medical image reports.
Specifically, Our model surpasses others in terms of CIDEr\cite{cider} and ROUGE-L\cite{rouge} metrics, while achieving comparable results in BLEU-4\cite{bleu} and METEOR metrics. The superior CIDEr values indicate that our model avoids redundant sentences from the training set and produces reports with more precise and relevant topics. The improvement in the CIDER metric indicates that our model has effectively addressed the issue of generating redundant text and to some extent mitigated data bias.
Additionally, we prioritize clinical correctness in our approach.

\subsubsection{Clinical Efficacy}
We evaluate the model by adopting chexbert\cite{chexbert} to extract common abnormalities from the generated radiology reports. Due to ChexBERT\cite{chexbert} was only trained on MIMIC-CXR, we present the clinical effectiveness metrics of MIMIC-CXR to demonstrate the clinical efficacy of our model. Since could not access the code of some methods, we only compare our results with some methods can be reproduced or report the clinical effectiveness. In \ref{table 2:clinical efficacy},it is evident that our model has made substantial advancements in terms of clinical effectiveness, exhibiting notable improvements in clinical accuracy, precision, and F1 score. These results highlight the remarkable efficacy of our model. Moreover, it demonstrates that incorporating prior abnormal knowledge as input effectively aids the model in focusing on abnormal information when generating corresponding medical imaging reports. This approach leads to the production of more reliable medical imaging reports.

\begin{table*}[t]
\caption{Comparative results of CAMANet with previous studies. The best values are highlighted in bold and the second best are underlined. BL and RG, MTOR and CIDR are the abbreviations of BLEU, ROUGE, METEOR and CIDEr respectively.}
\centering
\label{tab:main_results}

\begin{tabular}{clccccccc}
\toprule  
\textbf{Dataset}& \textbf{Method} &\textbf{BL-1} & \textbf{BL-2} & \textbf{BL-3} &\textbf{BL-4}  &\textbf{RG-L} & \textbf{MTOR} &\textbf{CIDR}  \\ 
\midrule  
\multirow{9}{*} {\textbf{IU-Xray}}&$ADAATT$ &0.220 &0.127 &0.089 &0.068 &0.308 & - & 0.295\\
\multirow{9}{*} &$ATT2IN$ &0.224 &0.129 &0.089 &0.068 &0.308 & - &0.220 \\
\multirow{9}{*} &$HRGR$ &0.438 &0.298 &0.208 &0.151 &0.322 & - &{0.343}\\
\multirow{9}{*} &$CoAT$ &0.455 &0.288 &0.205 &0.154 &0.369 & - &0.277 \\
\multirow{9}{*} &$CMAS-RL$ &0.464 &0.301 &0.210 &0.154 &0.362 & - &0.275\\
\multirow{9}{*} &$R2Gen$ &0.470 &0.304 &0.219 &0.165 &0.371 &0.187 &0.398\\ 
\multirow{9}{*} &$KERP$ &0.482 &0.325 &0.226 &0.162 &0.339 & - &0.280\\
\multirow{9}{*} &$CMCL$ &0.473 &0.305 &0.217 &0.162 &0.378 & 0.186\\
\multirow{9}{*} &$R2GenCMN^{*}$ &0.475 &0.309 &0.222
&0.170 &{0.375} & 0.191 &- \\
\multirow{9}{*} &$ARGG$ &0.475 &0.309 &0.222
&0.170 &{0.375} & 0.191 &- \\
\multirow{9}{*}  &$ (Ours)$  &\textbf{0.505} &\textbf{0.322} &\textbf{0.242} &\textbf{0.178}  &\textbf{0.382} & \textbf{0.193} &\textbf{0.454}\\
\midrule

\multirow{8}{*}{\textbf{MIMIC}} &$RATCHET$ &0.232 &- &- &- &0.240 & 0.101 &-\\
\multirow{8}{*}{\textbf{-CXR}} &$ST$ &0.299 &0.184 &0.121 &0.084 &0.263 & 0.124& - \\
\multirow{8}{*} &$ADAATT$ &0.299 &0.185 &0.124 &0.088 &0.266 & 0.118 &-\\
\multirow{8}{*} &$ATT2IN$ &0.325 &0.203 &0.136 &0.096 &0.276 & 0.134 &-\\
\multirow{8}{*} &$TopDown$ &0.317 &0.195 &0.130 &0.092 &0.267 & 0.128 &0.073 \\

\multirow{9}{*} &$CMCL$ &0.344 &0.217 &0.140 &0.097 &0.281 & 0.133 & -\\

\multirow{9}{*} &$R2Gen$ &0.353 &0.218 &0.145 &0.103 &0.277 & 0.142 &0.253\\ 
\multirow{5}{*} &$R2GenCMN$ &0.353 &0.218 &0.148 &0.106 &0.278 & 0.142 &-\\
\multirow{9}{*} &$ARGG$\cite{ARRG} &0.351 &0.223 &\textbf{0.157}
&\textbf{0.118} &\textbf{0.287} & -& \textbf{0.281} \\
\multirow{5}{*}  &$(Ours)$  &\textbf{0.373} &\textbf{0.227} &0.151 &0.107 &0.281 & \textbf{0.145}&0.264 \\

\bottomrule 
\end{tabular}
\end{table*}

\begin{table}[]
    \centering
    \begin{tabular}{c|ccc}
    \toprule
         \textbf{Methods}& \textbf{Precision} & \textbf{Recall} &\textbf{ F1-score} \\
    \midrule
    TopDown& 0.166 &0.121 &0.133 \\
    M2TR&0.240 & 0.428 &0.308\\
    Show-Tell&0.249&0.203&0.204\\
    
    R2Gen&0.333&0.273&0.276\\
    R2GenCMN&0.334 &0.275&0.278\\
    
    \midrule
   Ours&\textbf{0.377}&\textbf{0.312} &\textbf{0.315}\\
    \bottomrule 
    \end{tabular}
    \caption{The clinical efficacy metrics of our methods on MIMIC-CXR, comparing with other methods}
    \label{table 2:clinical efficacy}
\end{table}

\begin{table}[t]
\caption{The experimental results of component ablation studies, The best values are highlighted in bold.}
\centering
\label{tab:ablation_studies_component}
\begin{tabular}{l|ccccc}
\toprule  
\textbf{Models}  & \textbf{BL-3} & \textbf{RG-L} & \textbf{MTOR} &\textbf{F1-score}  \\
\midrule  

Ours   &0.151 &0.281 & 0.145& \textbf{0.315} \\
General Retrieval   &0.147 &0.273 & 0.138& 0.265 \\
w/o Retrieval   &0.138 &0.252 & 0.115& 0.243 \\
w/o FC mechanism &\textbf{0.165} & \textbf{0.293}& \textbf{0.152} & 0.275\\
\bottomrule
\end{tabular}
\end{table}

\subsection{Ablation Study}

\subsubsection{Effectiveness of every component}
To assess the effectiveness of each module, we conducted ablation experiments specifically targeting those modules. 
In order to contrast with the general approach of using image encoding vectors for retrieval, we trained the network's backbone using a commonly used image-based autoencoder for image reconstruction. Subsequently, we extracted image encoding vectors through the encoder for retrieval purposes.

In Table \ref{tab:ablation_studies_component}, the term "with general" denotes the traditional retrieval approach. Furthermore, we incorporated the anomalous sentences retrieved by this method as prior knowledge embeddings into the model. The metrics reveal no significant degradation in language generation quality, but there is a noticeable decline in clinical effectiveness. After applying the general retrieval method in the model, there was a significant decrease of 5\% in the F1 score. This suggests that our disease-oriented approach effectively enhances the quality of retrieved similar reports, thereby improving the model's perception of diseases.

"W/o Retrieval" refers to the model structure after removing the entire retrieval branch, causing the model to become a general seq2seq model that generates image reports solely through a transformer encoder-decoder. The model experienced a significant decrease in both language quality and clinical effectiveness. Specifically, language quality metrics such as BL-3, RG-L, and MTOR decreased by 1.3\%, 2.9\%, and 3.0\%, respectively. The F1 score showed a substantial decline of 7.2\%. These results indicate that embedding prior knowledge can effectively assist the model in generating higher quality medical image reports. Further exploration is warranted in the realm of more efficient knowledge integration.

"W/o FC mechanism" indicates the absence of a fact-consistent decoder based on the copying mechanism in the decoder. Similarly, we observed that the generated medical image reports did not exhibit significant degradation in language quality, but there was a severe decline in clinical effectiveness. Surprisingly, multiple language generation evaluation metrics of the model exhibited improvements, but there was a severe decline in clinical effectiveness indicators. Specifically, the language generation quality metrics, BL-3, RG-L, and MTOR, improved from 1.151 to 0.165, from 0.281 to 0.293, and from 0.145 to 0.152, respectively. However, the F1 score experienced a decline from 0.315 to 0.275.
The increase in language generation metrics may be attributed to the fact that when the model removes the copying mechanism, it tends to generate more normal descriptions. As normal descriptions constitute a large portion of medical image reports, it becomes a "shortcut" for improving language generation metrics. However, the decline in clinical effectiveness indicators demonstrates that although language generation metrics have improved, the actual quality of generated medical image reports has decreased. The introduced copying mechanism effectively utilizes prior input knowledge at the vocabulary level during the generation of medical image reports, leading the model to generate descriptions similar to the prior knowledge. This ultimately improves the clinical effectiveness of the model.

Our ablation experiments on multiple branches validate the effectiveness of the disease-oriented masked similar report retrieval and the fact-consistent decoder based on the copying mechanism proposed in our model.

\subsubsection{The amount of reference reports}
We also conducted an ablation experiment to investigate the effectiveness of using different numbers of reference image reports in generating medical image reports. We maintained the same hyperparameter settings, and the specific experimental results are shown in \ref{tab:ablation_studies_number}. It was observed that when three reference image reports were used as input, the model achieved the best performance. This phenomenon suggests that in the generation of image reports, an excessive number of reference reports can introduce redundant information, causing the model to overlook important information in the prior knowledge. On the other hand, using too few reference reports can result in insufficient experiential knowledge for the model, leading to performance degradation.

\begin{figure*}
    \centering
    
    \includegraphics[width=0.95\textwidth]{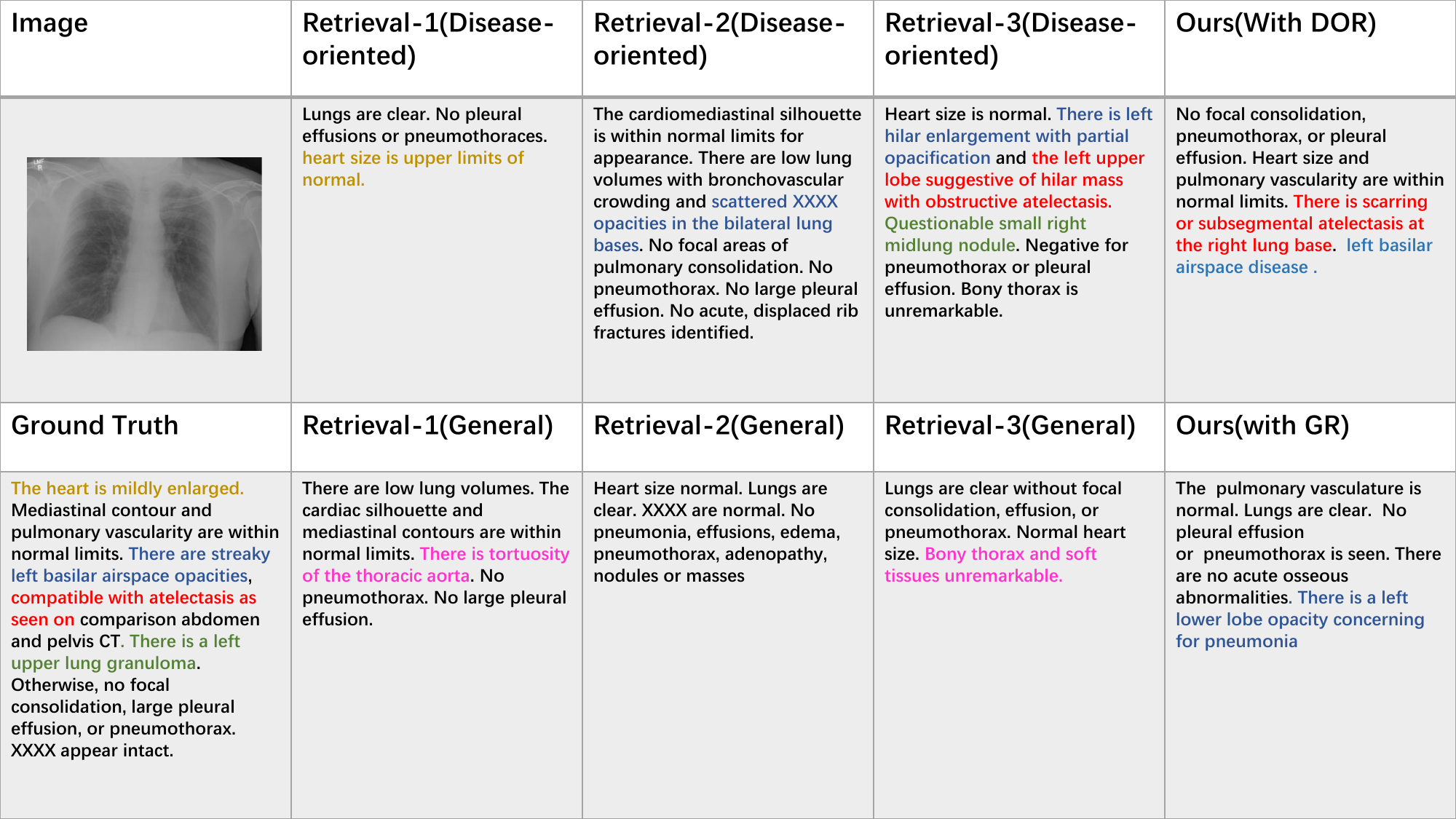}
    \caption{In an example from MIMIC-CXR, disease-related sentences in the image report are color-coded, which serves as the prior knowledge embedded into the model. By comparing two retrieval methods, it can be observed that the model retrieves more accurate medical image reports. Furthermore, from the generated results, it is evident that the generated reports exhibit a high degree of similarity with the retrieved report sentences, demonstrating the effectiveness of the fact-consistent generator.}
    \label{fig:different report}
\end{figure*}

\begin{figure*}
    \centering
    \includegraphics[width=0.95\textwidth]{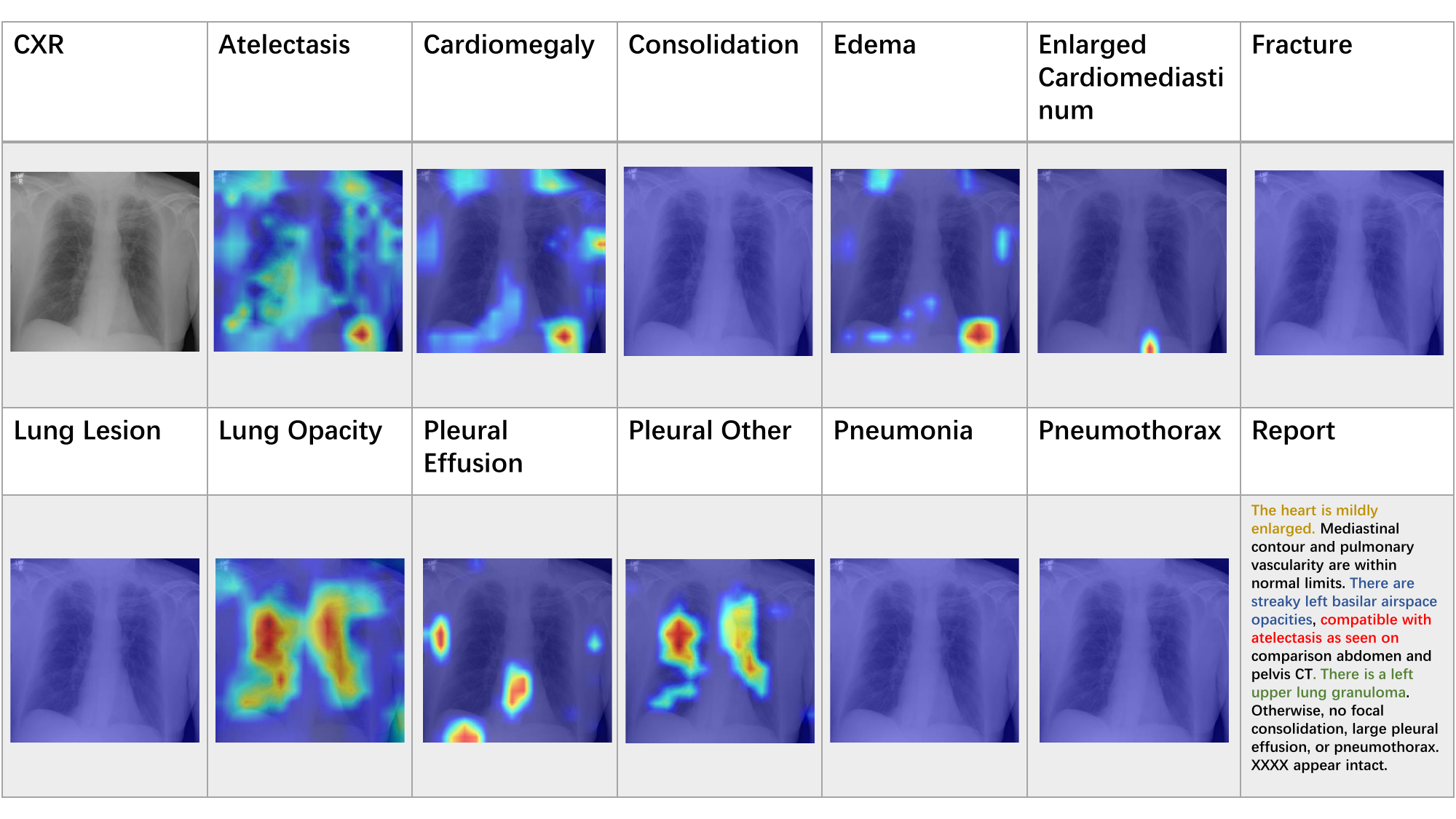}
    \caption{In an example from MIMIC-CXR, it can be observed that the CAM (Class Activation Map) effectively captures the disease's affected location and extent of information. The aggregated disease-oriented mask demonstrates a high level of semantic representation capability. As a result, it can be utilized to retrieve higher-quality medical image reports.}
    \label{fig:cam visual}
\end{figure*}

\subsection{Qualitative Results:} 

The \ref{fig:different report} presents a quantitative analysis of our model on the MIMIC dataset, where we provide a comparison between the reports generated by our model and the similar image reports retrieved through different methods. We have highlighted the abnormal sentences in different colors, while the reports describe normal conditions. We compare the reports obtained using the disease-oriented mask retrieval approach with those obtained using the conventional image retrieval method, selecting three reference reports as benchmarks. Through this comparison, we have found that the disease-oriented mask retrieval approach yields more accurate reference reports compared to the traditional image reconstruction-based image retrieval method. We observe that many disease-related abnormal description sentences in the Ground Truth exhibit a high degree of similarity with the abnormal description sentences in the retrieved reports, particularly in terms of the location information and morphological features of the diseases, indicating a significant overlap. By comparing the retrieved reports, the reports generated by our model, and the Ground Truth, we find that our proposed fact-consistent decoder based on the copy mechanism effectively incorporates relevant information from the reference similar reports. This enables the model to produce descriptive statements for important disease descriptions by leveraging the copied content, thereby enhancing the model's perception of diseases and clinical effectiveness in generating medical image reports.

\begin{table}[t]
\caption{The ablation study about the amount of reference reports , The best values are highlighted in bold.}
\centering
\label{tab:ablation_studies_number}
\begin{tabular}{l|cccc}
\toprule  
\textbf{Amount}  & \textbf{BL-3} & \textbf{RG-L} & \textbf{MTOR}&\textbf{F1-score} \\
\midrule  
1   &0.142 &0.265& 0.130 &0.275 \\
2   &0.146 &0.273 & 0.138 & 0.304 \\
3  & \textbf{0.151} &\textbf{0.281} &\textbf{0.145}&\textbf{0.315}\\
4   &0.154 &0.278 & 0.142&0.311\\
5 &0.149 &0.276 & 0.140&0.309 \\
\bottomrule
\end{tabular}

\end{table}

\section{Conclusion}
We propose a method based on disease-oriented mask retrieval, knowledge embedding, and fact-consistent image report generation.
Our work encompasses two main innovations. Firstly, we are the first to generate disease-oriented masks for each sample by utilizing a classification model to generate multi-class class activation maps and then aggregating and reducing them. These disease-oriented masks possess powerful disease representation capabilities, encompassing rich disease categories, morphology, and positional information. Extensive experimental results demonstrate that disease-oriented masks can replace original images for retrieval, as the retrieved medical image reports often exhibit high similarity to the target reports. This provides the model with strong prior knowledge, significantly enhancing its clinical effectiveness. Secondly, we propose a fact-consistent decoder based on the copy mechanism. This decoder can effectively leverage vocab-level prior information while also considering the input image information, enabling comprehensive output generation. Extensive experiments demonstrate that our model achieves remarkable performance improvements in terms of language generation quality and clinical effectiveness on two large benchmark datasets, IU-Xray and MIMIC-CXR. Moreover, the disease-oriented masks we propose can be further paired with medical image reports due to their stronger semantic representation capabilities.

The fact-consistent decoder based on the copy mechanism, proposed in our work, can be effectively applied to the domain of medical image report generation, which requires highly specialized training. Its powerful copying ability simulates the process of radiologists writing image reports, and qualitative analysis indicates that the decoder can replicate highly credible prior knowledge, thus enhancing the clinical effectiveness of our proposed model.

For future work, as our approach heavily relies on similar text reports as knowledge, which may have relatively narrow expertise, and lacks broad domain knowledge as credibility constraints, we look forward to incorporating medical textbooks like PUMed to improve the model's understanding of knowledge. Additionally, using widely accepted textbook language can provide factual constraints for image report generation models, expanding their applicability.

\bibliographystyle{IEEEtran}

\bibliography{ref.bib}
\end{document}